\documentclass[11pt]{article}

\usepackage[margin=1in]{geometry}
\usepackage{amsmath,amssymb}
\usepackage{booktabs}
\usepackage{graphicx}
\usepackage{microtype}
\usepackage[round,authoryear]{natbib}
\usepackage[hidelinks]{hyperref}
\usepackage{xcolor}
\usepackage{caption}

\graphicspath{{figures/}}
\newcommand{\EthicsStatement}{This retrospective study used de-identified drug-induced sleep endoscopy images and was approved by the institutional ethics committee (Approval No.~M2023504).}

\newcommand{\CompetingInterests}{The authors declare no competing interests.}

\title{ASTRA-Net: Anatomy-Specific Transfer and Representation Alignment for Drug-Induced Sleep Endoscopy Segmentation}

\author{
Suhua Sun$^{\dagger,1}$, Yuqiao Wang$^{\dagger,1,2,3}$, Sheng Liu$^{2}$, Rui Fan$^{1}$,\\
Jiajun Wang$^{2}$, Ruoyan Xu$^{2}$, Yixin Chen$^{2}$,\\
Tao Li$^{1}$, and Yan Yan$^{*,1}$\\[0.6em]
\small $^{1}$Department of Otolaryngology, Peking University Third Hospital, Beijing, China\\
\small $^{2}$Institute of Medical Technology, Peking University Health Science Center, Beijing, China\\
\small $^{3}$National Biomedical Imaging Center, College of Future Technology, Peking University, Beijing, China\\
\small $^{\dagger}$Equal contribution. $^{*}$Correspondence: \href{mailto:yanyan_ent@bjmu.edu.cn}{yanyan\_ent@bjmu.edu.cn}
}
\date{}

\begin{document}
\maketitle

\begin{abstract}
Quantitative drug-induced sleep endoscopy (DISE) requires reliable airway boundaries at specific anatomical levels. Pixel-level DISE annotations are scarce, and manual contouring limits the scalability of quantitative assessment. To address this limitation, we developed ASTRA-Net for known-plane DISE segmentation with limited real annotations. Stage 1 aligned intermediate ConvNeXt-Base representations from 14,250 unlabeled virtual endoscopy frames derived from computed tomography and real DISE frames. Virtual images were used only for feature alignment. Stage 2 fine-tuned four independent UNet++ decoders on 401 real annotated frames. Structured zero-mask supervision constrained incompatible plane outputs and invalid frames. Six alignment configurations used maximum mean discrepancy, domain adversarial learning, or both objectives. On a hold-out evaluation set of 100 frames, the five-model MMD-only segmentation ensemble achieved a mean Dice of 0.8927, with a 95\% image-level bootstrap interval of 0.8631 to 0.9160. The mean intersection over union was 0.8239. A classification-enabled variant of the same alignment configuration reached a restricted four-plane top-1 accuracy of 0.92 on the same hold-out frames. These results indicate that ASTRA-Net can support frame-level, plane-specific DISE boundary delineation when real annotations are limited.
\end{abstract}


\section{Introduction}

Obstructive sleep apnea is a common disorder of recurrent upper-airway collapse during sleep \citep{young1993occurrence}, and a literature-based analysis estimated that close to one billion adults worldwide are affected \citep{benjafield2019global}. Continuous positive airway pressure is the reference treatment \citep{sullivan1981reversal}, but patients who cannot tolerate it are considered for surgical or structural intervention, whose success depends on localizing the site and pattern of collapse. Drug-induced sleep endoscopy (DISE) provides dynamic visualization of upper-airway collapse in obstructive sleep apnea. Clinical interpretation is commonly organized by the velum, oropharynx, tongue base, and epiglottis (VOTE) classification \citep{kezirian2011vote}. Throughout this work, an anatomical plane denotes a VOTE-level label, where the velum corresponds to the soft palate, rather than a geometric imaging plane. Quantitative analysis requires a reproducible airway boundary at the assessed anatomical level. However, current computer-assisted measurements still rely on manual frame selection and delineation \citep{lai2020quantitative}, and visual interpretation remains observer dependent \citep{vroegop2013observer}. Cross-sectional area and orthogonal diameters are descriptive measures of luminal change after contour delineation \citep{lai2020quantitative}. A frame-level boundary at an identified anatomical level is therefore required before these quantities can be calculated. Plane-specific segmentation can supply a candidate boundary for such calibrated measurements at a known anatomical level. This work addresses that static, frame-level segmentation task and does not evaluate calibrated area, temporal stability, or clinical agreement.

Automatic DISE analysis remains limited by the visual conditions of endoscopy. Motion blur, specular reflection, secretions, and occlusion obscure the airway boundary. Nonrigid deformation changes the visible lumen within and across anatomical levels. Annotation requires reviewers to distinguish the lumen margin from collapsed soft tissue under variable illumination. The target boundary can be absent when a frame does not show a valid view of the assessed plane. Against these conditions, recent learning-based studies predicted obstruction scores at the clip or examination level \citep{hanif2023automatic,kim2026automatic}. These categorical outputs do not provide pixel boundaries for continuous area or diameter measurements. Dense boundary annotation is expensive, so the number of real DISE masks is limited.

Virtual endoscopy derived from computed tomography (CT) can extend the geometric and anatomical view diversity available for model development. Exposing a model to broad synthetic variation is a common strategy for narrowing the gap to real data \citep{tobin2017domain}. Synthetic endoscopic datasets support image analysis through controllable anatomical views and simulator-derived task annotations \citep{martyniak2025simuscope}. Endoscopic lesion adaptation has transferred segmentation knowledge from labeled source images to unlabeled target images \citep{dong2020transfer}. The present setting differs in supervision and target definition. It uses CT-derived upper-airway views without segmentation masks, plane labels, or image-level correspondence with clinical DISE. Real DISE masks define the visible airway lumen at a specified VOTE level. CT-derived airway renderings preserve gross airway geometry. Virtual views can sample geometric variation without adding clinical contouring work. Their appearance does not reproduce the illumination, secretions, motion, and dynamic tissue deformation encountered during examination. This synthetic-to-real discrepancy limits direct use of virtual images for clinical segmentation.

Domain-adaptation methods address image-distribution differences through source-domain task labels, image translation, pseudo-labels, or feature alignment \citep{kamnitsas2017adaptation,chen2019sifa,zhang2024mapseg}. Source-domain labels define the segmentation semantics in many of these settings. Objectives that depend on source masks cannot define a virtual-domain segmentation target in the present setting. Pseudo-labeling and image translation would require task-consistent pseudo-labels, translation fidelity, or semantic-preservation properties that cannot be verified with the available virtual data. Real DISE masks define the segmentation target in the present study. Virtual views contribute visual and geometric information during representation learning. The learning process uses cross-domain information without assigning segmentation semantics to unlabeled virtual images. It retains the anatomical specificity required for VOTE-level assessment. Existing DISE studies address manual measurements or categorical video assessment \citep{lai2020quantitative,hanif2023automatic,kim2026automatic}, and neither supplies plane-specific pixel boundaries. The present setting therefore combines three properties that the studies cited above do not jointly satisfy: the task semantics are the VOTE-level airway lumen, the source domain provides no virtual mask or plane label, and the target domain offers only limited real masks under known-plane evaluation. This unlabeled-virtual, real-mask-supervised setting for plane-specific DISE segmentation motivates the present study.

\begin{figure}[t]
  \centering
  \includegraphics[width=\textwidth]{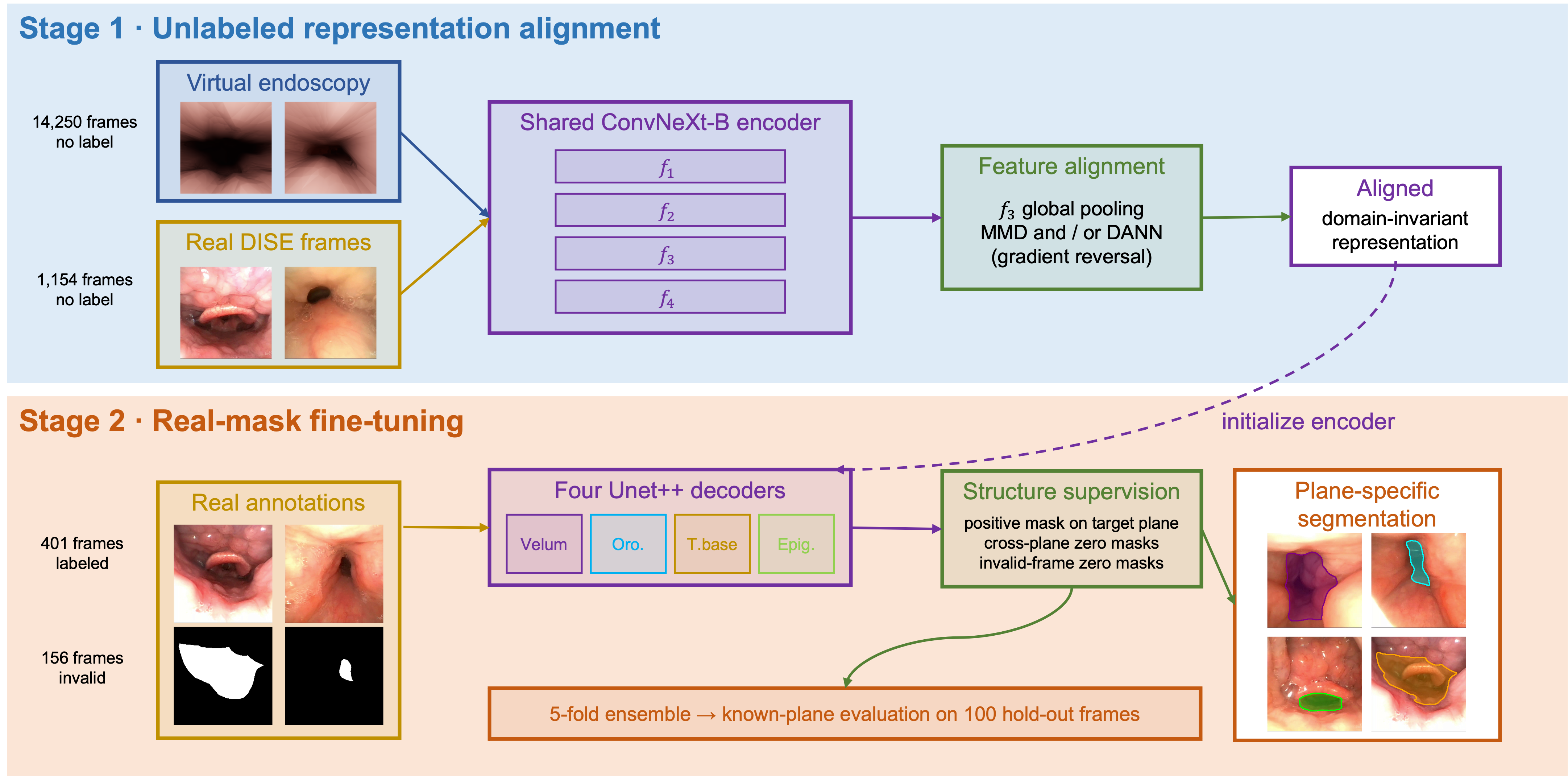}
  \caption{Study overview. CT-derived virtual airway views extend the image domain available for representation learning. Real DISE annotations define plane-specific segmentation and support evaluation on held-out real DISE frames.}
  \label{fig:framework}
\end{figure}

This work investigates plane-specific DISE segmentation with limited real masks and unlabeled CT-derived virtual views. Figure~\ref{fig:framework} presents the study design. The evaluated task is static known-plane segmentation; automatic plane routing, invalid-frame rejection, temporal analysis, and clinical quantification are outside the present scope. The contributions are as follows.
\begin{enumerate}
  \item We formulate an anatomy-specific transfer setting for DISE in which unlabeled CT-derived virtual views enter representation alignment without being assigned segmentation semantics, while real masks remain the sole segmentation supervision for the VOTE-level airway boundary. Whether the virtual views improve segmentation is left to future real-only comparisons.
  \item We introduce ASTRA-Net for this setting and implement a segmentation pipeline that aligns virtual and real representations before real-mask training, and that uses anatomy-specific decoders with structured zero-mask supervision for incompatible planes and invalid-frame inputs, without evaluating invalid-frame rejection.
  \item We report hold-out analyses of known-plane segmentation and, as an auxiliary measure, restricted four-class anatomical-plane classification on valid real DISE frames. This classification is a top-1 label analysis over the four VOTE levels and does not constitute automatic routing or a clinical-workflow evaluation. The analyses are a within-pipeline comparison of six evaluated feature-alignment configurations with paired image-level differences, and they do not attribute effects to individual components.
\end{enumerate}

\section{Related Work}

\subsection{Clinical interpretation and quantitative DISE}

Drug-induced sleep endoscopy is interpreted through the anatomical level, degree, and configuration of upper-airway collapse. The European position paper defined a shared procedure and reporting language \citep{devito2018position}. The velum, oropharynx, tongue base, and epiglottis classification provides the anatomical vocabulary used in clinical assessment \citep{kezirian2011vote}. Agreement varies across anatomical levels and reader experience \citep{kezirian2010interrater,vroegop2013observer,mitsikas2025interrater}. Reproducible image boundaries could complement ordinal assessment with continuous geometric measurements.

Lai et al. measured cross-sectional area and orthogonal diameters after manual delineation at four upper-airway levels \citep{lai2020quantitative}. Deep learning studies have predicted obstruction grades from clips or examinations \citep{hanif2023automatic,kim2026automatic}. These approaches address measurement after manual contouring or categorical interpretation. Plane-specific pixel segmentation supplies a frame-level boundary that is necessary, but not sufficient, for automated geometric analysis, which further requires spatial calibration, temporal stability, and clinical validation beyond the known-plane scope of this study.

\subsection{Medical and endoscopic image segmentation}

Encoder--decoder architectures with multiscale skip pathways, such as U-Net and UNet++, and hierarchical convolutional encoders, such as ConvNeXt, supply transferable dense-prediction backbones for medical image segmentation \citep{ronneberger2015unet,zhou2018unetpp,liu2022convnet}. Self-configuring pipelines and transformer-based encoders extend this family and set strong benchmarks across segmentation tasks \citep{isensee2021nnunet,chen2021transunet}. These designs address the extraction of multiscale boundary features and are adopted here as the encoder and decoder substrate. They do not, however, specify the output semantics when the target is a level-dependent structure: the segmentation target is defined only after an anatomical level and a valid view are established. Endoscopic segmentation methods further target weak and ambiguous boundaries under reflection, fluid, blur, and irregular illumination; PraNet addressed such boundary ambiguity in colonoscopy \citep{fan2020pranet}. This line transfers to the appearance conditions of DISE but assumes a single, always-present target structure. DISE segmentation instead targets the visible airway lumen at a specified VOTE level, so the plane label and the presence of a valid frame, rather than the backbone alone, determine what the output should represent. Because dense annotation is costly, medical segmentation under limited labels has been approached through semi-supervised consistency training and self-supervised few-shot learning \citep{tarvainen2017meanteacher,ouyang2020superpixel}. These methods reduce annotation on the target domain itself; the present study instead keeps real masks as the only segmentation supervision and draws additional, label-free variation from an unlabeled virtual domain.

\subsection{Synthetic data and domain adaptation}

Synthetic endoscopy can increase the diversity of training images. SimuScope combines surgical simulation with appearance translation and produces simulator-derived annotations \citep{martyniak2025simuscope}. It is useful to separate the kinds of variation such data provide: geometric and anatomical variation, appearance variation, and clinical realism are distinct. The CT-derived virtual upper-airway images in this study contribute anatomical and geometric variation only; they have no masks, plane labels, or image-level correspondence with DISE, and their rendered appearance does not reproduce clinical endoscopic conditions. They enter feature alignment only, and their inclusion is not evidence that synthetic data improve segmentation, which the present design does not test.

Learning useful representations without task labels is well established through self-supervised contrastive and masked-image objectives \citep{he2020moco,chen2020simclr,he2022mae}, which motivates using unlabeled images to shape a shared feature space. Feature-distribution alignment provides objectives in this spirit that do not require paired data: maximum mean discrepancy compares distributions through kernel mean embeddings \citep{gretton2012kernel}, and domain adversarial neural networks use gradient reversal to reduce domain information in learned features \citep{ganin2016domain}. Medical unsupervised domain adaptation, in contrast, commonly propagates the segmentation target from a labeled source domain through source masks, cycle-consistent image translation, or pseudo-labels \citep{zhu2017cyclegan,kamnitsas2017adaptation,chen2019sifa,zhang2024mapseg}. Comparing these mechanisms along a common set of axes clarifies the present gap: whether the virtual (source) domain carries masks or plane labels, whether a cross-domain semantic correspondence is assumed, the form of real-domain supervision, and whether the output is a VOTE-level pixel boundary. Among the representative mechanisms compared here, the virtual domain has no mask or plane label and no reliable semantic correspondence with DISE, so source-mask, translation, and pseudo-label objectives cannot define the segmentation target; the label-free distribution-alignment objectives are the ones that remain directly usable for cross-domain representation learning in this setting. ASTRA-Net therefore uses unlabeled virtual images for representation alignment and defines the segmentation target solely from real masks, without a virtual segmentation target, pseudo-label, or image translation. Video segmentation methods additionally use sequence information and temporal representations \citep{ji2021pnsnet,ji2022video,li2022tccnet}; ASTRA-Net is evaluated as a spatial model and reports no temporal metric.

\begin{table}[t]
\centering
\caption{Positioning of representative methods and the present study.}
\label{tab:related_work}
\small
\setlength{\tabcolsep}{3.5pt}
\begin{tabular}{p{0.13\textwidth}p{0.18\textwidth}p{0.19\textwidth}p{0.13\textwidth}p{0.11\textwidth}p{0.14\textwidth}}
\toprule
Research line & Representative work & Primary supervision & Temporal information & Plane-specific pixel output & Relation to this study \\
\midrule
DISE interpretation & VOTE and position paper \citep{kezirian2011vote,devito2018position} & Expert ordinal scores & Reviewed clinically, not modeled & No & Defines the anatomical levels and reporting context \\
Quantitative DISE & Lai et al. \citep{lai2020quantitative} & Manual geometric measurements & Manual frame selection & Manual & Establishes measurements that require a reproducible boundary \\
Automated DISE scoring & Hanif et al. and Kim et al. \citep{hanif2023automatic,kim2026automatic} & Clip or examination scores & Aggregated over video & No & Predicts categorical collapse and provides no pixel boundaries \\
Endoscopic segmentation & PraNet \citep{fan2020pranet} & Real pixel masks & Not evaluated (framewise) & Single target & Addresses weak boundaries in another endoscopic anatomy \\
Medical UDA & Kamnitsas et al., SIFA, and MAPSeg \citep{kamnitsas2017adaptation,chen2019sifa,zhang2024mapseg} & Typically source masks with translation or pseudo-labels & Usually not evaluated & Source-defined & Propagates a source-defined target to an unlabeled domain \\
Synthetic endoscopy & SimuScope \citep{martyniak2025simuscope} & Simulator-derived annotations (virtual task labels) & Simulated procedure data & From virtual labels & Uses virtual images that carry task labels \\
ASTRA-Net & Present study & Real masks; virtual images unlabeled & Not evaluated (framewise) & Yes (VOTE level) & Aligns intermediate features before plane-specific training \\
\bottomrule
\end{tabular}
\end{table}

The literature establishes the clinical need for quantitative DISE boundaries and the available mechanisms for segmentation and domain adaptation. Relative to the representative works positioned in Table~\ref{tab:related_work}, ASTRA-Net occupies the combination of an unlabeled virtual source domain, real-mask supervision, and a VOTE-level pixel output; we do not claim priority over the broader literature. The reported comparisons concern six alignment configurations within the same pipeline and use known-plane frame-level evaluation.

\section{Method}

\subsection{Problem formulation and learning stages}

Let $x^v$ and $x^r$ denote virtual and real DISE images. Each annotated real image has a binary airway mask $y$ and one of four anatomical planes $p\in\mathcal{P}$, where $\mathcal{P}$ contains soft palate, oropharynx, tongue base, and epiglottis. A separate collection of real frames supplies a five-dimensional multi-hot label, comprising the four plane labels and an invalid-frame label, for the classification head. The static evaluation reports a restricted four-plane decision on valid frames.

All images were resized to $256\times256$ pixels and normalized with ImageNet statistics. Stage 1 aligned encoder features from unlabeled virtual and real images. Stage 2 initialized the encoder from Stage 1 and trained the plane-specific decoders with real annotations. Virtual masks, virtual plane labels, and image-level correspondence were absent from both training objectives.

\subsection{Encoder and unlabeled feature alignment}

The virtual dataset comprised 14,250 RGB views rendered from 44 CT-derived upper-airway surface models. An ImageNet-1K-initialized ConvNeXt-Base encoder produced multiscale features. The third encoder level supplied pooled features for domain alignment. The remaining levels supplied spatial features for segmentation and global features for plane classification.

Alignment operated on the third-level features. For each image the third encoder level was globally average pooled to a $512$-dimensional vector, and the pooled virtual and real vectors of a mini-batch were standardized per dimension over their joint statistics. Let $z_i^v$ and $z_j^r$ denote these standardized virtual and real features, with $n_v$ and $n_r$ vectors per domain. The alignment used a multi-kernel Gaussian discrepancy, with $k_{\beta}(u,w)=\exp(-\lVert u-w\rVert^2/(2\beta^2))$ and $k$ the average of the kernels at bandwidths $\beta\in\{1,5,10\}$. MMD minimized the empirical discrepancy
\begin{align}
\mathcal{L}_{\mathrm{MMD}}={}&\frac{1}{n_v^2}\sum_{i,i'}k(z_i^v,z_{i'}^v)
+\frac{1}{n_r^2}\sum_{j,j'}k(z_j^r,z_{j'}^r) \nonumber\\
&-\frac{2}{n_vn_r}\sum_{i,j}k(z_i^v,z_j^r),
\end{align}
where $k$ is the arithmetic mean of the three Gaussian kernels defined above. DANN used a gradient reversal layer (GRL) and a domain classifier $D_\phi$ applied to the same pooled features. For domain label $d_i$ (0 for virtual, 1 for real) and classifier logit $a_i$, the loss was
\begin{equation}
\mathcal{L}_{\mathrm{DANN}}
=-\frac{1}{n_v+n_r}\sum_i
\left[d_i\log\sigma(a_i)+(1-d_i)\log(1-\sigma(a_i))\right],
\qquad a_i=D_\phi(\operatorname{GRL}_\gamma(z_i)).
\end{equation}
The domain classifier $D_\phi$ was a compact nonlinear head mapping the pooled feature to a single domain logit. The gradient reversal scaling $\gamma$ increased with training progress $t\in[0,1]$ following the schedule $\gamma=\min(2t,1)$, so that the adversarial signal was weak early in training and reached full strength in the second half. The Stage 1 objective was
\begin{equation}
\mathcal{L}_{\mathrm{pre}}=\lambda_m\mathcal{L}_{\mathrm{MMD}}+\lambda_d\mathcal{L}_{\mathrm{DANN}}.
\end{equation}
The six configurations varied the weights $(\lambda_m,\lambda_d)$ over the $50$ Stage 1 epochs, as listed in Table~\ref{tab:configs}: MMD-only $(1,0)$, DANN-only $(0,1)$, a fixed mixture $(0.8,0.2)$, and three schedules that ramp the two weights across three epoch phases (the first $30\%$, the middle $40\%$, and the final $30\%$ of epochs). Segmentation and plane-classification losses were disabled during Stage 1, so virtual images affected the encoder representation only.

\begin{table}[t]
\centering
\caption{Stage 1 feature-alignment configurations. Scheduled settings give $(\lambda_m,\lambda_d)$ over three epoch phases; fixed settings apply throughout.}
\label{tab:configs}
\small
\begin{tabular}{lll}
\toprule
Setting & Alignment record & $(\lambda_m,\lambda_d)$ over epochs\\
\midrule
C1 & Scheduled DANN+MMD & $(0.8,0.2)\!\to\!(0.5,0.5)\!\to\!(0.2,0.8)$\\
C2 & DANN only & $(0,1)$ throughout\\
C3 & MMD only & $(1,0)$ throughout\\
C4 & DANN-to-MMD & three-phase ramp, $\lambda_m$ increasing\\
C5 & MMD-to-DANN & three-phase ramp, $\lambda_d$ increasing\\
C6 & Fixed mixture & $(0.8,0.2)$ throughout\\
\bottomrule
\end{tabular}
\par\vspace{2pt}
\footnotesize The scheduled settings (C1, C4, C5) each vary $(\lambda_m,\lambda_d)$ across three epoch phases spanning the first $30\%$, the middle $40\%$, and the final $30\%$ of the $50$ epochs. They differ in the direction of the ramp between the two objectives; C1 is stated explicitly above, and C4 and C5 are distinct scheduled variants.
\end{table}

\subsection{Plane-specific segmentation}

The segmentation network used four plane-specific UNet++ decoders on shared multiscale encoder features, one per anatomical plane, each producing a binary airway probability map. The decoders did not share parameters. Each could then specialize to the lumen definition of a single VOTE level, whose appearance and shape differ across levels, and the plane-specific negative supervision described below applied to an independent output. Shared and plane-conditioned decoders define alternative designs that are not evaluated here. A separate five-way classifier on the global encoder feature predicted the four plane labels and an invalid-frame label under a multi-label formulation, trained with a binary cross-entropy loss over the multi-hot label $\mathbf{c}\in\{0,1\}^5$:
\begin{equation}
\mathcal{L}_{\mathrm{cls}}=-\frac{1}{5}\sum_{l=1}^{5}\left[\mathbf{c}_l\log\sigma(a_l)+(1-\mathbf{c}_l)\log(1-\sigma(a_l))\right].
\end{equation}
The classification and segmentation objectives share the encoder and are combined with a weight $\lambda_{\mathrm{cls}}$ on the classification term, so that $\lambda_{\mathrm{cls}}$ governs how strongly plane classification shapes the shared representation. The reported segmentation metrics correspond to $\lambda_{\mathrm{cls}}=0$, for which the encoder is driven by the segmentation objective alone, and the classification results in Section~\ref{sec:experiments} correspond to a positive classification weight. Segmentation was evaluated in known-plane mode, which selects the reference-plane decoder and therefore excludes plane-routing errors.

\subsection{Structured negative supervision}

For a predicted probability map $\hat{y}$ and binary target $y$ over pixels $u$, the segmentation loss combined soft Dice and focal terms. The soft Dice term was
\begin{equation}
\mathcal{L}_{\mathrm{Dice}}(\hat{y},y)
=1-\frac{2\sum_u \hat{y}_uy_u+\epsilon}{\sum_u \hat{y}_u+\sum_u y_u+\epsilon},
\end{equation}
with a small smoothing constant $\epsilon>0$ that also defines the loss as zero for an all-zero prediction against an all-zero (negative-supervision) target.
The focal term was the pixel-mean binary focal loss
\begin{equation}
\mathcal{L}_{\mathrm{Focal}}(\hat{y},y)
=-\frac{1}{|U|}\sum_u \alpha_u\,(1-q_u)^{\varsigma}\log q_u,
\end{equation}
where $q_u=\hat{y}_u$ if $y_u=1$ and $q_u=1-\hat{y}_u$ otherwise, $\alpha_u=0.75$ for foreground and $0.25$ for background pixels, the focusing exponent is $\varsigma=2$, and $|U|$ is the number of pixels. The combined loss used unit weights
\begin{equation}
\ell(\hat{y},y)=\mathcal{L}_{\mathrm{Dice}}(\hat{y},y)+\mathcal{L}_{\mathrm{Focal}}(\hat{y},y).
\end{equation}

The target-plane decoder received the annotated mask. Selected off-plane decoders received an all-zero target. The negative-head mapping was
\begin{equation}
\mathcal{N}(p)=
\begin{cases}
\{\mathrm{O},\mathrm{S}\}, & p\in\{\mathrm{T},\mathrm{E}\},\\
\{\mathrm{T},\mathrm{E},\mathrm{S}\}, & p=\mathrm{O},\\
\{\mathrm{T},\mathrm{E},\mathrm{O}\}, & p=\mathrm{S},
\end{cases}
\end{equation}
where $\mathrm{T}$, $\mathrm{E}$, $\mathrm{O}$, and $\mathrm{S}$ denote tongue base, epiglottis, oropharynx, and soft palate. This mapping is fixed from anatomy rather than learned: oropharynx and soft palate are penalized against the other levels because their lumen views are visually distinct, whereas tongue base and epiglottis receive no mutual penalty because those structures can co-occur near anatomical transitions.

For an annotated mini-batch $\mathcal{B}$ and an accompanying set $\mathcal{I}$ of invalid frames, the segmentation objective was
\begin{align}
\mathcal{L}_{\mathrm{fine}}=\frac{1}{Z}\bigg[&
\sum_{i\in\mathcal{B}}\ell(\hat{y}_{i,p_i},y_i)
+\rho\sum_{i\in\mathcal{B}}\sum_{h\in\mathcal{N}(p_i)}
\ell(\hat{y}_{i,h},\mathbf{0}) \nonumber\\
&+\eta\sum_{r\in\mathcal{I}}\sum_{h\in\mathcal{P}}
\ell(\hat{y}_{r,h},\mathbf{0})\bigg],
\end{align}
where the first term supervises the target-plane decoder with the annotated mask, the second applies zero-mask targets to the incompatible decoders $\mathcal{N}(p_i)$, and the third applies zero-mask targets to all four decoders on invalid frames. The weights were $\rho=1$ for the incompatible-plane term and $\eta=0.15$ for the invalid-frame term, and $Z$ is the number of accumulated terms, so the objective is their mean. Invalid frames were included alongside the annotated frames at a small, fixed ratio. The weights $\rho$ and $\eta$ were fixed rather than tuned, and the present experiments include no sensitivity or ablation analysis of the mapping or these weights. This structured supervision constrained both the incompatible plane outputs and all plane outputs on invalid frames.

\subsection{Ensemble and known-plane inference}

The five fold-specific models were combined into a single predictor by averaging their outputs. Denoting the number of models by $K=5$, the plane scores and the plane-specific mask probabilities were averaged separately,
\begin{equation}
\bar{a}=\frac{1}{K}\sum_{k=1}^{K}a^{(k)}, \qquad
\bar{y}_p=\frac{1}{K}\sum_{k=1}^{K}\hat{y}^{(k)}_p,
\end{equation}
with the plane scores averaged before the logistic transform and the mask probabilities after it. Restricted four-plane top-1 accuracy used the largest of the four plane entries of $\sigma(\bar{a})$. Known-plane evaluation selects the reference-plane probability map, thresholds it at $0.5$, and refines the resulting mask with the operator $\mathcal{R}$,
\begin{equation}
m_p=\mathcal{R}\left(\mathbb{I}[\bar{y}_p\geq0.5]\right),
\end{equation}
where $\mathcal{R}$ fills interior holes of the thresholded mask and then retains the largest connected component, additionally keeping the second-largest component only when its area is at least one fifth of the largest. Connectivity is four-connected for the background holes and eight-connected for the foreground components. Optional video utilities are described in Appendix~\ref{app:video} and lie outside the frame-level evaluation.

\subsection{Training and model-selection protocol}

Both stages were trained with AdamW under cosine learning-rate annealing and a weight decay of $10^{-4}$. Stage 1 ran for 50 epochs with encoder and domain-classifier learning rates of $5\times10^{-6}$ and $5\times10^{-5}$, the smaller encoder rate letting the alignment signal adapt the pretrained encoder gradually. The single aligned encoder from Stage 1 initialized every Stage 2 fold. Stage 1 used only the plane-labeled and invalid real frames for alignment; the 100 hold-out frames were excluded from both stages, so the evaluation is not transductive. Stage 2 trained each of the five folds for 40 epochs with encoder and decoder learning rates of $10^{-5}$ and $10^{-4}$ and with the domain classifier held fixed. For each fold, the parameters at the epoch of highest mean validation Dice were selected.

\section{Experiments}
\label{sec:experiments}

\subsection{Data inventory and evaluation design}

\begin{figure}[t]
  \centering
  \includegraphics[width=\textwidth]{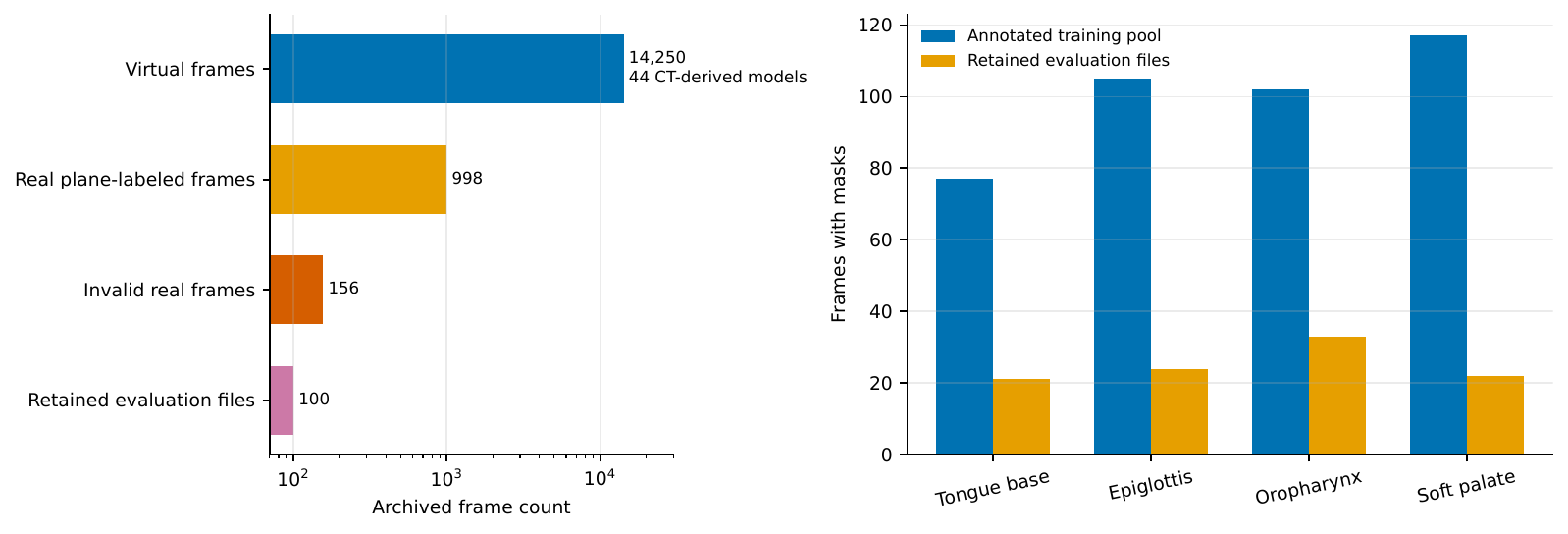}
  \caption{Frame inventory for training and hold-out evaluation. The virtual images are unlabeled. The 44 CT-derived entries supplied virtual views. All reported counts are frame counts.}
  \label{fig:data}
\end{figure}

The study used 998 real frames with plane labels, 401 real frames with segmentation masks, 156 invalid frames, and 14,250 virtual frames from 44 CT-derived models. The masked training collection contained 77 tongue-base frames, 105 epiglottic frames, 102 oropharyngeal frames, and 117 soft-palate frames. The hold-out evaluation set contained 100 frames, including 21 tongue-base, 24 epiglottic, 33 oropharyngeal, and 22 soft-palate masks, as shown in Figure~\ref{fig:data}. All reported counts are frame counts.

The available annotations are frame-level, and subject, examination, and video identifiers were not accessible for the real frames. The sampling unit is therefore the frame, and independence at the subject or video level cannot be verified; the 100 hold-out frames were held out as whole frames. This limits the analysis to frame-level generalization and is revisited in the limitations. The masked, plane-labeled, and invalid collections were treated as distinct frame sets for their respective supervision roles, and the 100 hold-out frames were disjoint from every training frame.

Stage 1 used unlabeled virtual and real frames. Stage 2 used real masks for segmentation supervision. Plane-labeled real frames supplied classifier supervision. Invalid real frames supplied zero-mask targets. The 401 masked frames were randomly partitioned, under a fixed seed for reproducibility, into five approximately equal folds of 80 or 81 frames. Each of the five models was trained on four folds and used the held-in fifth fold as its internal validation subset, from which the epoch with the highest mean validation Dice was kept. The five resulting models were combined into a single ensemble by averaging their outputs, as defined in the ensemble inference above. This cross-validation partition was applied only to the 401 training frames. The 100 hold-out frames formed a separate set that entered no fold and was used for neither training, model selection, nor threshold selection; the segmentation threshold was fixed at 0.5 throughout. The same hold-out set was evaluated for all six alignment configurations, which enabled paired image-level comparisons.

The comparison is internal to the pipeline: the six alignment configurations of Table~\ref{tab:configs} are the compared conditions, the primary endpoint is the mean known-plane Dice on the hold-out frames, and the fixed-mixture configuration serves as the reference condition for the paired differences. The design contains no real-only (alignment-free) condition, no external segmentation baseline, and no component-removal ablation. The comparison therefore isolates the choice of alignment objective, not the contribution of alignment, virtual data, or structured supervision in itself. The remaining metrics and all non-reference comparisons are descriptive.

\subsection{Metrics and statistical analysis}

For each binarized prediction, true positives $T_P$, true negatives $T_N$, false positives $F_P$, and false negatives $F_N$ were counted over all pixels. The five frame-level metrics were
\begin{align}
\mathrm{Dice} &= \frac{2T_P}{2T_P+F_P+F_N+\epsilon}, &
\mathrm{IoU} &= \frac{T_P}{T_P+F_P+F_N+\epsilon},\\
\mathrm{Precision} &= \frac{T_P}{T_P+F_P+\epsilon}, &
\mathrm{Recall} &= \frac{T_P}{T_P+F_N+\epsilon},\\
\mathrm{Specificity} &= \frac{T_N}{T_N+F_P+\epsilon}, &&
\end{align}
where $\epsilon$ is a small constant that only guards against division by zero; because every hold-out frame has a non-empty reference mask, no empty-target case arises, and a frame with an empty prediction against a non-empty target scores zero Dice. Restricted four-plane top-1 accuracy was computed over the 100 hold-out frames, each of which carries one of the four anatomical reference planes and none of which is invalid. For each frame the predicted plane was the largest of the four anatomical entries of the ensemble-averaged plane probabilities, with the invalid entry excluded from this comparison, and accuracy was the fraction of frames whose predicted plane matched the reference plane.

Overall segmentation values are image-weighted means. For a metric $m_i$ calculated on frame $i$, the reported estimate is
\begin{equation}
\bar{m}=\frac{1}{N}\sum_{i=1}^{N}m_i, \qquad N=100.
\end{equation}
Plane-specific values used the same calculation within each anatomical subset. Mean Dice is the primary endpoint and carries bootstrap intervals; IoU, precision, recall, specificity, and restricted plane accuracy are reported as descriptive secondary measures without intervals.

Percentile 95\% confidence intervals (CIs) for mean Dice used 10,000 bootstrap resamples under a fixed seed. If $I_b=(i_1,\ldots,i_N)$ is a sample of image indices drawn with replacement, bootstrap replicate $b$ is
\begin{equation}
\bar{D}^{(b)}=\frac{1}{N}\sum_{i\in I_b}D_i.
\end{equation}
The interval endpoints were the 2.5th and 97.5th percentiles of $\{\bar{D}^{(b)}\}$, and the point estimate was the observed mean Dice. All alignment settings shared the same image identifiers. For the reference setting $c$ and a comparator $a$, each bootstrap replicate reused the same drawn indices $I_b$ for both settings, preserving the per-image pairing
\begin{equation}
\Delta_{a,c}^{(b)}=\frac{1}{N}\sum_{i\in I_b}(D_{i,a}-D_{i,c}).
\end{equation}
The paired-difference interval used the same 2.5th and 97.5th percentiles of $\{\Delta_{a,c}^{(b)}\}$, with the observed mean difference as the point estimate. The paired comparisons against the reference are descriptive and exploratory; no multiple-comparison correction was applied, and an interval that excludes or includes zero is reported as such rather than as a confirmed superiority claim. Because subject and video identifiers were unavailable, resampling was performed over frames; if several frames originate from one subject or video, this frame-level resampling can understate the true intervals, and the intervals should be read as frame-level. The per-image metric records supporting these calculations are being prepared for public release.

\section{Results}

\subsection{Alignment configurations}

\begin{figure}[t]
  \centering
  \includegraphics[width=\textwidth]{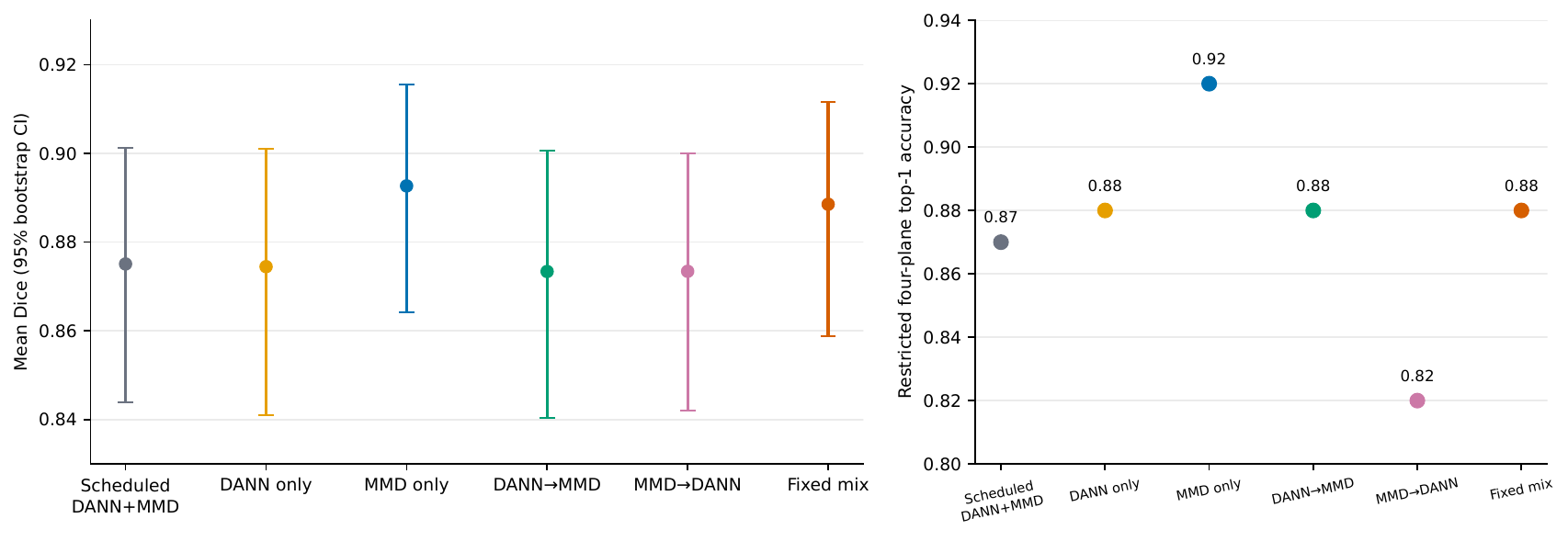}
  \caption{Known-plane mean Dice with 95\% image-level bootstrap intervals and restricted four-plane top-1 accuracy for six alignment configurations.}
  \label{fig:alignment}
\end{figure}

\begin{figure}[t]
  \centering
  \includegraphics[width=0.78\textwidth]{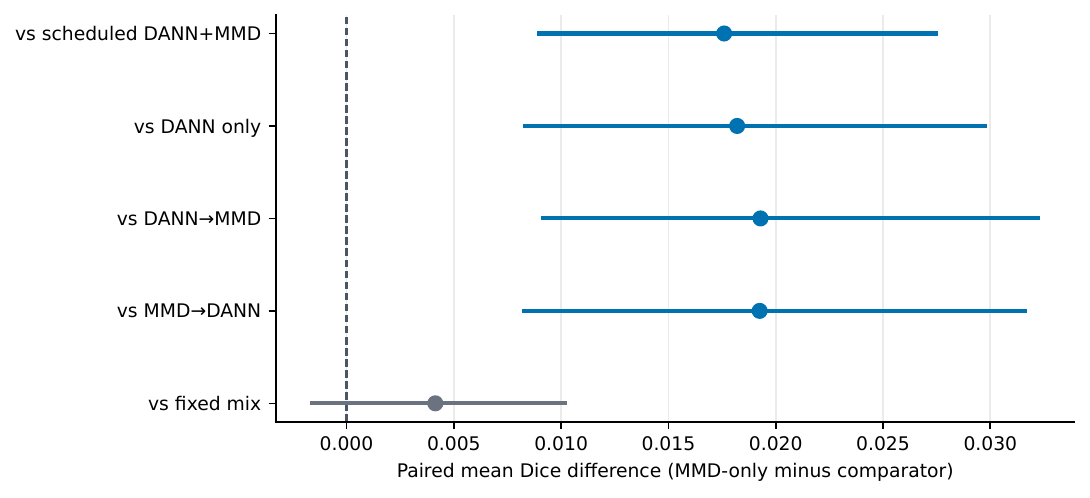}
  \caption{Paired mean Dice differences between MMD-only and each comparator. Positive values favor MMD-only. The interval against the fixed mixture includes zero.}
  \label{fig:paired}
\end{figure}

Table~\ref{tab:alignment} and Figure~\ref{fig:alignment} compare six Stage 1 configurations within the same pipeline. The Dice and IoU values are from segmentation-only ensembles ($\lambda_{\mathrm{cls}}=0$); the restricted plane accuracy values are from classification-enabled variants of each configuration and are reported alongside for completeness but are not from the same trained models. MMD-only achieved the highest mean Dice and restricted four-plane accuracy. Its paired Dice difference from the fixed mixture was 0.0041. The image-level interval included zero, as shown in Figure~\ref{fig:paired}.

\begin{table}[t]
\centering
\caption{Performance on the hold-out set of 100 frames. Segmentation uses the known plane. Dice and IoU are from the segmentation-only ensemble ($\lambda_{\mathrm{cls}}=0$); restricted plane accuracy is from a classification-enabled variant of each configuration. Intervals are 95\% image-level bootstrap intervals.}
\label{tab:alignment}
\small
\begin{tabular}{lllll}
\toprule
Setting & Alignment record & Dice (95\% CI) & IoU & Restricted plane accuracy\\
\midrule
C1 & Scheduled DANN+MMD & 0.8751 (0.8431--0.9011) & 0.7991 & 0.87\\
C2 & DANN only & 0.8745 (0.8422--0.9014) & 0.7991 & 0.88\\
C3 & MMD only & 0.8927 (0.8631--0.9160) & 0.8239 & 0.92\\
C4 & DANN-to-MMD & 0.8734 (0.8407--0.9004) & 0.7977 & 0.88\\
C5 & MMD-to-DANN & 0.8734 (0.8420--0.8995) & 0.7969 & 0.82\\
C6 & Fixed mixture & 0.8885 (0.8595--0.9114) & 0.8173 & 0.88\\
\bottomrule
\end{tabular}
\end{table}

The remaining four configurations had lower mean Dice values than MMD-only and the fixed mixture. Their restricted four-plane accuracies ranged from 0.82 to 0.88. The paired intervals in Figure~\ref{fig:paired} quantify differences on the shared hold-out frames.

\subsection{Plane-specific segmentation}

The MMD-only configuration (C3) achieved the highest mean Dice on the hold-out set among the six compared configurations. The following plane-level characterisation is therefore exploratory: C3 was identified from the same 100-frame hold-out set used for reporting, so the values below should not be read as pre-specified confirmatory outcomes. The MMD-only ensemble reached a mean Dice of 0.8927, as detailed by plane in Table~\ref{tab:plane}. Mean precision was 0.9018. Mean recall was 0.9064. Mean specificity was 0.9910. Tongue base and oropharynx achieved higher mean Dice values than soft palate and epiglottis. Among the four planes, the epiglottis, evaluated on 24 frames, had the widest image-level interval.

\begin{table}[t]
\centering
\caption{Plane-specific known-plane performance of the MMD-only ensemble on the hold-out frames. Intervals are 95\% image-level bootstrap intervals. Values are not end-to-end routing performance.}
\label{tab:plane}
\small
\begin{tabular}{lrrrrrr}
\toprule
Plane & $n$ & Dice (95\% CI) & IoU & Precision & Recall & Specificity\\
\midrule
Tongue base & 21 & 0.9424 (0.9260--0.9558) & 0.8931 & 0.9497 & 0.9401 & 0.9824\\
Epiglottis & 24 & 0.8329 (0.7316--0.9107) & 0.7587 & 0.8323 & 0.8800 & 0.9889\\
Oropharynx & 33 & 0.9201 (0.9071--0.9324) & 0.8541 & 0.9271 & 0.9193 & 0.9974\\
Soft palate & 22 & 0.8693 (0.8125--0.9097) & 0.7838 & 0.8939 & 0.8838 & 0.9920\\
\midrule
Overall & 100 & 0.8927 (0.8631--0.9160) & 0.8239 & 0.9018 & 0.9064 & 0.9910\\
\bottomrule
\end{tabular}
\end{table}

\begin{figure}[htbp]
  \centering
  \includegraphics[width=0.96\textwidth]{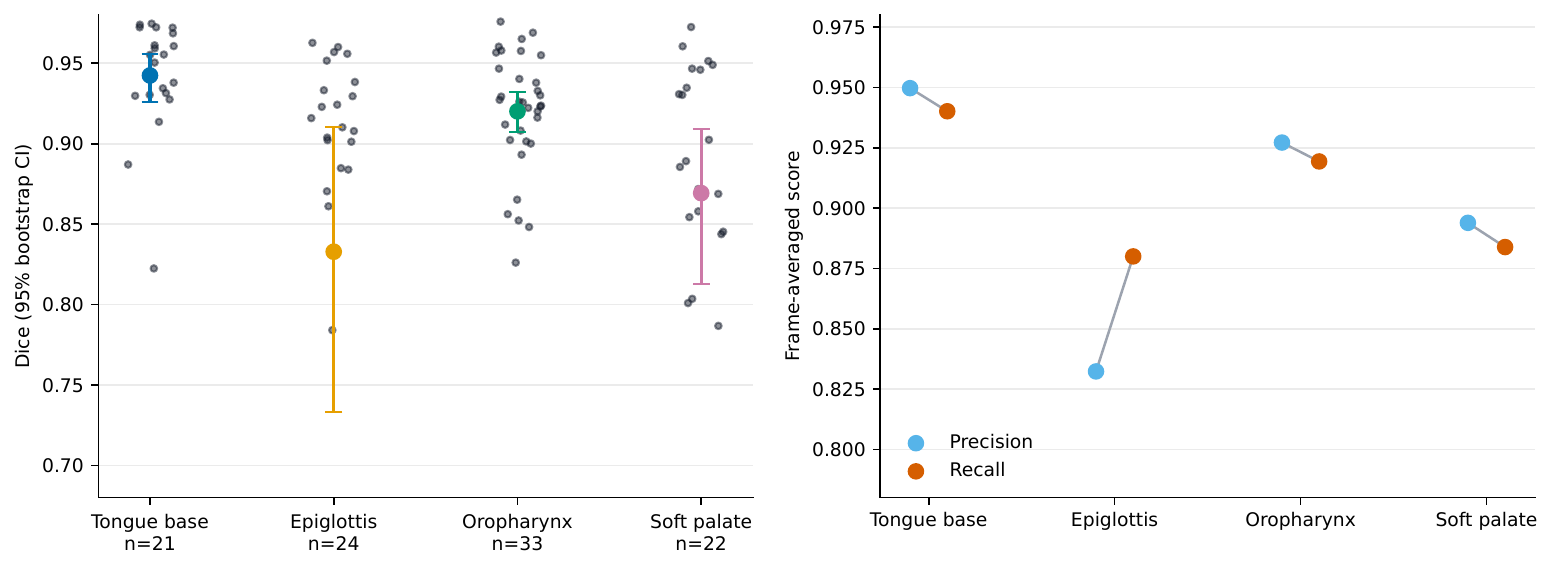}
  \caption{Known-plane segmentation by anatomical level. Gray points show Dice for individual frames. Colored points and error bars show the means and 95\% image-level bootstrap intervals. Precision and recall are averaged across frames and displayed as paired points.}
  \label{fig:plane}
\end{figure}

Precision and recall had similar overall means, and their per-plane values are shown in Figure~\ref{fig:plane}. The plane subsets contained between 21 and 33 frames. Restricted plane classification was evaluated separately from the known-plane segmentation task.

\subsection{Qualitative error analysis}

\begin{figure}[htbp]
  \centering
  \includegraphics[width=0.96\textwidth]{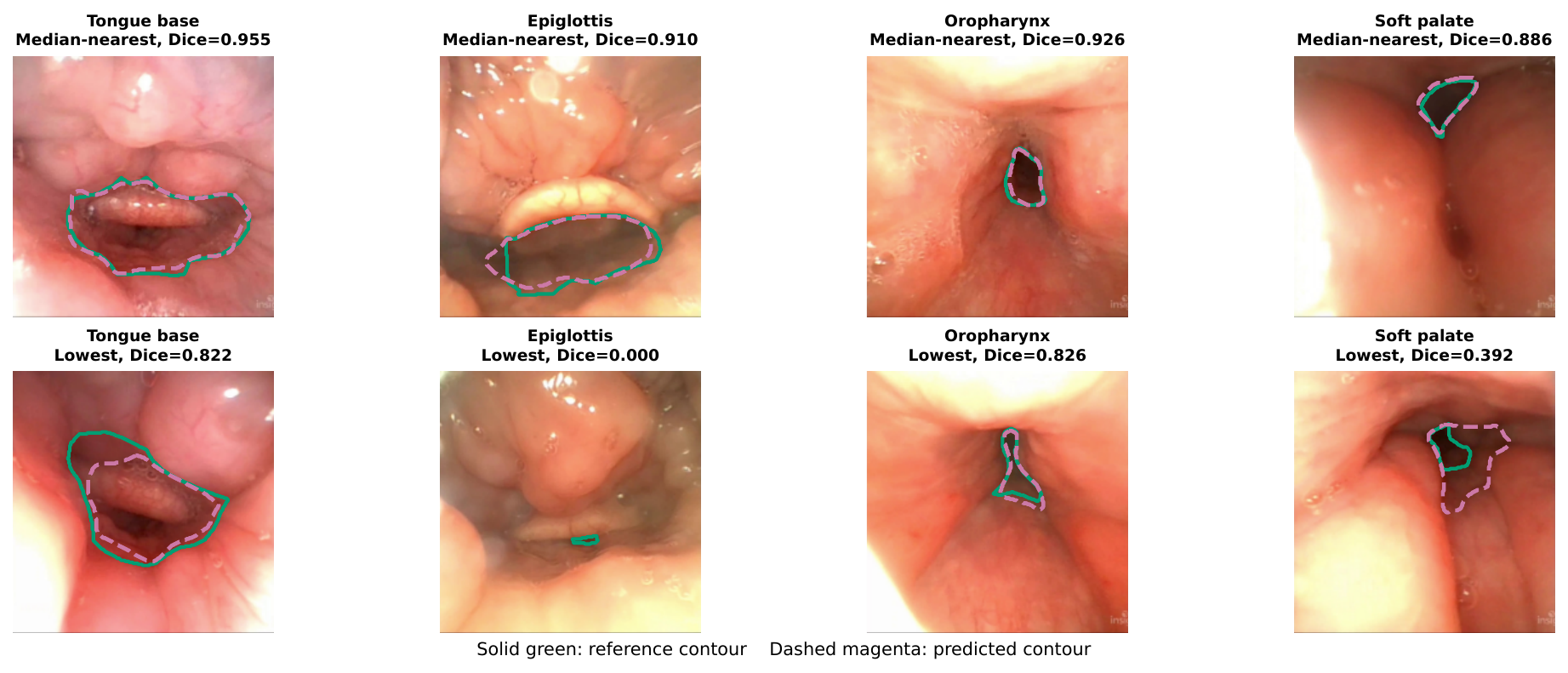}
  \caption{Reference contours are solid green and predicted contours are dashed magenta. Each plane includes a median-nearest example and a lowest-Dice example. The saved per-image metrics determined the selection. All identifiers were removed. The epiglottic failure has an empty prediction and no magenta contour.}
  \label{fig:qualitative}
\end{figure}

Figure~\ref{fig:qualitative} presents the median-nearest and lowest-Dice outputs for each anatomical plane. A fixed rule selected the examples without manual curation. Median-nearest examples show close contours. Low-Dice examples show undersegmentation or localization failure.

The median-nearest examples retained a visible predicted contour for all four planes. The lowest-Dice epiglottic example had an empty prediction. Other low-Dice examples showed disagreement in boundary extent or location.

\section{Discussion}

\subsection{Domain alignment and supervision}

The MMD-only configuration achieved the highest mean Dice among the six configurations. MMD directly constrains feature distributions through a batchwise discrepancy \citep{gretton2012kernel}, whereas DANN relies on adversarial optimization through a domain classifier \citep{ganin2016domain}. One unverified explanation is that, because the virtual and real image domains differ substantially in appearance, the direct MMD objective provided a more stable optimization signal in these runs. We did not measure training stability or the underlying optimization behavior, so this remains a hypothesis rather than a finding. The paired interval against the fixed mixture included zero, which supports a cautious interpretation of the observed ranking.

ASTRA-Net differs from medical adaptation pipelines that use image translation, pseudo-labels, or source-domain segmentation targets \citep{chen2019sifa,zhang2024mapseg}. Virtual images contributed representation alignment only. Real DISE masks defined the segmentation task. Plane-specific decoders represented the changing lumen definition across anatomical levels. Structured zero-mask targets constrained incompatible plane outputs and invalid frames. The present experiments evaluate the complete design, while matched ablations are required to quantify the contribution of each component.

Automated DISE scoring studies predict clip-level or examination-level obstruction grades \citep{hanif2023automatic,kim2026automatic}. Manual DISE analysis can provide quantitative measurements after contour delineation \citep{lai2020quantitative}. ASTRA-Net provides a pixel-level airway boundary for a selected anatomical plane. This separation of plane selection and boundary delineation follows the clinical workflow in which an anatomical level is identified before geometric assessment.

\subsection{Prospective quantitative use}

The predicted boundary can support future quantitative analysis. For a binary mask $m_t$ at frame $t$ and calibrated pixel dimensions $s_x$ and $s_y$, cross-sectional area and equivalent circular diameter are
\begin{equation}
A_t=s_xs_y\sum_u m_{t,u}, \qquad
d_{\mathrm{eq},t}=2\sqrt{\frac{A_t}{\pi}}.
\end{equation}
Without spatial calibration, these quantities remain in pixel units. Calibration is required for physical measurements.

A reference frame with area $A_{\mathrm{ref}}$ permits a fractional collapse measure
\begin{equation}
C_t=1-\frac{A_t}{A_{\mathrm{ref}}}.
\end{equation}
Framewise segmentation supplies the geometric object required by these calculations. Clinical interpretation also requires stable boundaries across respiratory motion. A complete automatic system will require plane identification, invalid-frame rejection, segmentation, and temporal analysis. The present study evaluates the known-plane segmentation component.

\subsection{Limitations}

The folds and hold-out evaluation use frame-level splits. Patient and video identifiers are unavailable, therefore patient-level independence could not be assessed. The evaluation set contains 100 frames from one archive and no external-center cohort. Future studies should use patient-grouped splits and multi-center validation.

The experiments compare alignment configurations within the same pipeline. They contain no real-only baseline, matched external segmentation baseline, or component-removal experiment. These studies are required to estimate the effects of virtual alignment and structured zero-mask supervision.

The study evaluates static known-plane segmentation. It does not evaluate automatic routing, invalid-frame rejection, temporal stability, collapse measurement accuracy, or clinical agreement. Dense video annotation and calibrated measurement protocols are needed for these outcomes.

\section{Conclusion}

This study presented ASTRA-Net for plane-specific DISE segmentation with limited real annotations. The method aligned features from unlabeled virtual endoscopy images and real DISE images before supervised training with real masks. Structured zero-mask supervision constrained incompatible plane outputs and invalid frames. The MMD-only ensemble achieved a mean known-plane Dice of 0.8927 on the hold-out set of 100 frames. Tongue base and oropharynx had the highest mean Dice values.

Among the six alignment configurations compared in this study, the MMD-only setting attained the highest descriptive hold-out performance; because the design includes no real-only condition, no component ablation, and no external baseline, these results do not isolate the contribution of feature alignment itself. Real-only baselines and component ablations are needed to estimate that contribution, and patient-grouped and multi-center validation, matched baselines, and temporal video evaluation are required before clinical deployment.

\section*{Data and Code Availability}

The source code, CT-derived virtual endoscopy images used for feature alignment, and de-identified per-image evaluation records are not publicly available at the time of submission because the materials are still being organized and documented for release. These materials will be made available in a public repository upon formal publication of this article. The real clinical DISE images, their corresponding annotations, and trained model weights are not publicly shared because of participant-privacy and institutional data-governance restrictions.

\section*{Ethics Statement}

\EthicsStatement

\section*{Competing Interests}

\CompetingInterests

\section*{Reproducibility Statement}

The source code, CT-derived virtual endoscopy images, and de-identified per-image evaluation records needed to reproduce the reported figures and statistical analyses are being prepared for public release upon formal publication. Full retraining will remain restricted because the real clinical images and annotations cannot be publicly shared, and trained model weights will not be distributed.

\appendix
\section{Claim and Evidence Boundary}

Table~\ref{tab:boundary} records the interpretation of the main outputs. The boundary limits the known-plane, frame-level results to the evaluation design of this study.

\begin{table}[h]
\centering
\caption{Interpretation of the reported evidence.}
\label{tab:boundary}
\small
\begin{tabular}{p{0.24\textwidth}p{0.31\textwidth}p{0.34\textwidth}}
\toprule
Output & Supported interpretation & Outside the evidence scope\\
\midrule
Mean Dice on 100 frames & Known-plane postprocessed frame segmentation & Patient-independent or automatic performance\\
Restricted plane accuracy & Four-plane top-1 classification on valid frames & Calibrated routing, invalid-frame rejection, or clinical scoring\\
Six alignment settings & Descriptive comparison of alignment objectives & Superiority over real-only or standard baselines\\
Optional video utility & Implemented causal postprocessing & Quantitatively validated temporal stability\\
Virtual images & Unlabeled feature alignment input & Virtual masks, anatomical labels, or learned anatomical prior\\
\bottomrule
\end{tabular}
\end{table}

\section{Unevaluated Video Utilities}
\label{app:video}

For completeness we sketch an optional, unevaluated causal smoothing rule for video input. It is a candidate for future work rather than a formalized component of the evaluated method. The idea is to modulate temporal smoothing by prediction confidence: with $q_{t,(1)}$ and $q_{t,(2)}$ the two largest anatomical probabilities at frame $t$, a margin-based uncertainty $u_t=1-(q_{t,(1)}-q_{t,(2)})$ increases when the top two planes are close, and a confidence-weighted running average
\begin{equation}
\tilde{q}_t=\alpha_t q_t+(1-\alpha_t)\tilde{q}_{t-1}
\end{equation}
smooths uncertain frames more strongly toward the previous estimate, with the blending weight $\alpha_t$ decreasing in $u_t$ and kept within a bounded range. A complementary rule would retain previous-frame pixels only near the current mask, and known-plane mode would fix the requested plane and disable the update. The rule carries no learned temporal state. The present study reports no temporal metric and makes no tracking claim, and the smoothing coefficients are left unspecified because this rule is outside the evaluated method.

\bibliographystyle{plainnat}
\bibliography{references}

\end{document}